\documentclass[10pt,twocolumn,letterpaper]{article}

\usepackage{wacv}
\usepackage{times}
\usepackage{epsfig}
\usepackage{graphicx}
\usepackage{amsmath}
\usepackage{amssymb}
\usepackage{authblk}

\wacvfinalcopy 

\def\wacvPaperID{****} 

\ifwacvfinal\fi
\begin{document}

\title{Future Semantic Segmentation Using 3D Structure}


\author[1,*]{Suhani Vora}
\author[1,2]{Reza Mahjourian}
\author[1]{Soeren Pirk}
\author[1]{Anelia Angelova}

\affil[1]{Google Brain, Mountain View CA 94043, USA. (svora,rezama,pirk,anelia)@google.com}
\affil[2]{University of Texas at Austin, Austin TX 78712, USA.}
\affil[*]{Work done as a member of the Google AI Residency program (g.co/airesidency).}
\setcounter{Maxaffil}{0}
\renewcommand\Affilfont{\itshape\small}

\maketitle
\ifwacvfinal\thispagestyle{empty}\fi

\begin{abstract}
Predicting the future to anticipate the outcome of events and actions is a critical attribute of autonomous agents; particularly for agents which must rely heavily on real time visual data for decision making. Working towards this capability, we address the task of predicting future frame segmentation from a stream of monocular video by leveraging the 3D structure of the scene. Our framework is based on learnable sub-modules capable of predicting pixel-wise scene semantic labels, depth, and camera ego-motion of adjacent frames. We further propose a recurrent neural network based model capable of predicting future ego-motion trajectory as a function of a series of past ego-motion steps. Ultimately, we observe that leveraging 3D structure in the model facilitates successful prediction, achieving state of the art accuracy in future semantic segmentation.
\end{abstract}

\section{Introduction}

An important aspect of intelligent behaviour for autonomous agents, is the ability to not only comprehend singular static scenes but also to leverage previously collected visual information, to predict the outcome of future events~\cite{DBLP:journals/corr/Shalev-ShwartzB16}. As a proxy for this capability, many attempts at forecasting future video from past frames have been made, centering around prediction of RGB pixel values~\cite{DBLP:journals/corr/KalchbrennerOSD16, DBLP:journals/corr/MathieuCL15, DBLP:journals/corr/RanzatoSBMCC14, DBLP:journals/corr/SrivastavaMS15}. However, prediction in RGB pixel space quickly leads to blurring effects as RGB pixel intensities are difficult to predict precisely ~\cite{7995953}. \par  

Accurate RGB prediction is also arguably unnecessary for many tasks.
An alternative approach has emerged centering around future prediction in the semantic segmentation space which contains sufficient detail of contextual information for decision making in many applications. Future semantic segmentation aims to predict pixel-wise semantic labels of future frames given a stream of past monocular RGB video frames. Prior work on future semantic segmentation focuses on directly mapping inputs to the future frame in an unstructured manner. In one approach, autoregressive CNNs take a stack of past frames and predict future frame segmentations directly, performing well against a segmentation baseline, and demonstrating convincing future frames up to 0.5 seconds in advance~\cite{8237339}. In another approach, Generative Adversarial Networks (GANs) are applied to produce temporally consistent features~\cite{8237857}, which complement and enhance a segmentation model's accuracy, by using both future and past frames.

\begin{figure}[t]
\begin{center}
  \includegraphics[width=\linewidth]{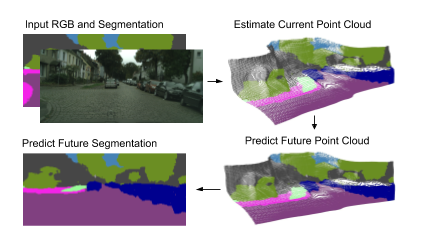}
  \caption{A high-level overview of our method.  The input RGB frame, along with a segmentation mask predicted from it, are used to construct a 3D point cloud (via a learnable depth estimation). This point cloud is then transformed according to a prediction for future egomotion trajectory. Finally,  projecting the predicted point cloud into 2D space produces a segmentation prediction for the future without access to the future RGB frame or motion data.}
  \label{fig:teaser}
\end{center}
\end{figure}

In this work, we propose a novel, and complementary to prior work, approach for future segmentation. 
We cast the segmentation problem into the 3D scene itself and propose to incorporate the structure information into learning (Figure~\ref{fig:teaser}). More specifically we propose learning each structured component of the scene: particularly the 3D geometry and the future ego-motion trajectory. Given a video sequence, we treat the future prediction as a result of a rigid transformation within a 3D scene. We apply a joint model to estimate 3D depth and ego-motion of past RGB frames,  to reliably transform past semantic segmentation masks to predict the segmentation of future frames. We treat ego-motion as an SE3 transform\cite{Byravan17} and further propose to train a recurrent neural network module to predict future frame ego-motion trajectory, which additionally enhances the accuracy of our predictions. 
To the best of our knowledge, 3D information such as depth and ego-motion have not been previously employed for this task, and our work is the first to introduce 3D structure as well as learned future motion to future semantic prediction. We demonstrate the usefulness of the approach by forecasting the segmentation of future frame sequences and are able to achieve improvements in segmentation accuracy on unseen future frames, relative to a segmentation-only baseline as well as a state-of-the-art result. The advantages of our method are: \par

\begin{itemize}
    \item Predicts future frames by means of 3D geometry information. One innovation of this work is to leverage 3D depth and ego-motion sub-modules for the purposes of future segmentation. These are learned in an unsupervised method from video sequence only, meaning no additional supervision or sensors are necessary.
    \item Treats the problem by introducing proper structure (3D geometry of the scene), thus allowing for clear paths to improvement by enhancing well understood sub-module i.e. improving ego-motion estimates by learning predicted SE3 transforms for future ego-motion trajectories.
    \item Provides competitive performance to the state-of-the-art and is capable of predicting future segmentations at larger time horizons.
\end{itemize}
 
\section{Related Work}

The first future frame prediction models were employed for directly predicting the next or future RGB video frames~\cite{DBLP:journals/corr/KalchbrennerOSD16, DBLP:journals/corr/MathieuCL15, DBLP:journals/corr/RanzatoSBMCC14, DBLP:journals/corr/SrivastavaMS15}. These methods work at the pixel level, entirely in the RGB space and lack semantics. These approaches are appealing as they only require RGB video frames for supervision, at the same time they often lead to blurred next frame predictions due to the uncertainty of each pixel's precise value and location. 

Alternatively, prediction of semantically labelled pixels greatly simplifies the task, as pixel values are categorical and uncertainties more easily resolved. Furthermore, employing semantic abstraction modules has been theoretically shown to lower the amount of training data required when modeling complicated tasks \cite{DBLP:journals/corr/Shalev-ShwartzS16}. \par \par

Future frame segmentation was introduced by a video scene parsing method by Luc et al.\cite{8237339}, in which the primary task was pixel-wise semantic labeling of future frames by learning and using features, which enforce temporal consistency. This was accomplished with a network, similar to~\cite{ DBLP:journals/corr/MathieuCL15}, that was trained to predict the immediate next frame from a series of four preceding frames. Luc et al.~\cite{8237339} further introduced an autoregressive convolutional neural network that is trained to iteratively generate multiple future frames. The model performs well relative to a baseline segmentation, as well as warping. In this prior work all training and evaluation is done on local image patches of 128x256 pixels in order to take advantage of the full image resolution. Conversely, in our work we aim to use the whole scene information instead and approach the task by incorporating 3D information for semantic understanding.

Jin et al~\cite{8237857} propose leveraging temporal consistency to enhance the segmentation task itself; future frames were used jointly with past frames, to produce the highest quality segmentation possible. Unlike this work we focus on producing future frame segmentation masks without employing future frame information. \par

More loosely related to our work, future frame prediction which decouples motion and content has been proposed by Villegas et al~\cite{Villegas17}. Video sequences with instance segmentations have also been successfully used for prediction of human or object trajectories for activity understanding~\cite{ 6909657, 6248012, Kitani:2012:AF:2404742.2404759, 10.1007/978-3-319-10578-9_45, 6126279}. \par

\begin{figure*}[t]
  \includegraphics[width=\linewidth]{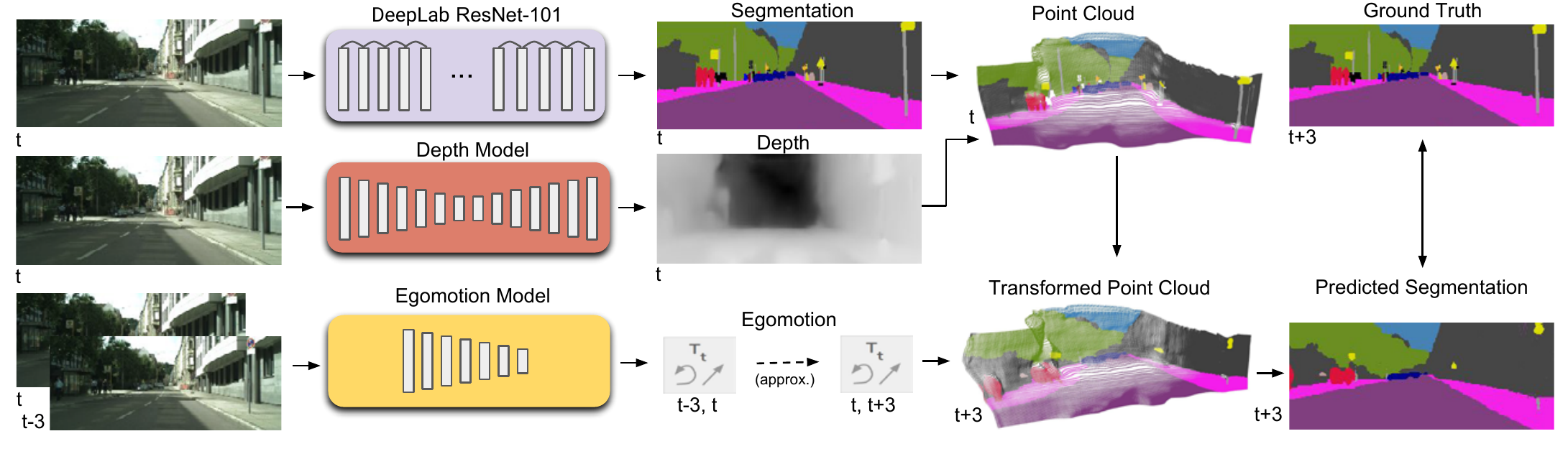} 
    \vspace{-4mm} 
  \caption{An overview of our method: to predict the segmentation of a future frame ($t+3$), we first compute the segmentation of the current frame ($t$) and estimate its depth (which is also learned). Together this allows us to generate a segmented 3D point cloud of the current frame ($t$). Additionally, we use the current frame ($t$) and its preceding frames to estimate the ego-motion of the future frames ($t+3$), ($t+6$), etc. We then use the predicted ego-motion to transform the segmented point cloud from frame ($t$) to generate the segmentation of the future frame ($t+3$), and recursively to subsequent frames ($t+6$), ($t+9$), etc. }
  \label{fig:overview}    
\end{figure*}
\vspace{-5pt}

\section{3D Future Semantic Segmentation}
The task of future scene segmentation is defined as follows.  Given input RGB images $X_{t-9}, X_{t-6}, X_{t-3}, X_{t}$ for frames $t-9$ to $t$, predict a segmentation map $S_{t+3j}$ for future frames $t+3j$, for $j \in \{1,2,3,4,5\}$, when available. We consider different versions of the problem by predicting three frames ahead of time (a short-term prediction~\cite{8237339} which is 0.18 seconds in the future) or 6-9 frames in advance (mid-term prediction which is up to 0.54 seconds), and longer term, i.e. 12-15 frames (up to 0.9 seconds in the future).
Each segmentation map $S$ is specified as a $K-$channel one-hot map $S^{ijk}$ where $K$ is the number of classes ($K=19$ for Cityscapes~\cite{Cityscapes}), $k \in \{1..K\}$, and $i, j$ iterate over the image coordinates.  At times, we drop the $k$ superscript and use $S^{ij}$ to refer to the entire one-hot vector of size $K$ at coordinates $i, j$.  We use a time subscript $t$, e.g. $S^{ij}_t$, to refer to the segmentation map at time $t$.

An overview of the 3D future semantic segmentation method and its components is shown in Figure~\ref{fig:overview}. In order to predict the segmentation of a future frame ($t+3$), we propose to reconstruct the scene in 3D.  We first compute the segmentation of the current frame at time $t$, e.g. by~\cite{journals/corr/Chollet16a,Chen2018-pv} and learn to predict its depth, which can be done without supervision~\cite{8100183,mahjourian2018unsupervised}. A segmented 3D point cloud of the current frame ($t$) is generated, which represents the 3D scene as observed.
We further propose to learn to predict the future ego-motion trajectory from the history of prior ego-motion vectors and apply this future egomotion to transform the 3D scene accordingly.
The 3D point cloud is then recurrently transformed to future positions according to the future ego-motion.
We note that future ego-motion trajectories are not available as ground truth and we use the learned ones from the unsupervised depth and ego-motion model. If ground truth ego-motion vectors are available through global positioning systems (GPS) they can significantly improve the future estimations as well.

This section is organized as follows:
Section~\ref{sec:base} describes the baseline segmentation model.
Section~\ref{ssec:3DSegmentationTransformation} presents the main 3D future segmentation, where also the strategy for learning the depth and ego-motion is described,
Section~\ref{ssec:Learn Future Ego-Motion} presents the proposed future ego-motion learning and prediction, and finally, Section~\ref{sec:inpaint} describes the inpainting approach which is part of the future segmentation.

\subsection{Baseline Segmentation Model}
\label{sec:base}

We employ a baseline semantic segmentation model based on Xception-65 \cite{journals/corr/Chollet16a} (or alternatively a model based on DeepLab ResNet-101 \cite{Chen2018-pv,7780459} can be used). The segmentation model produces a segmentation map  ${S}_{t}^{ij}$ with a 19 dimension one-hot class label vector (corresponding to the semantic classes available in Cityscapes), at each pixel ${i,j}$ in the frame. The input to this model is the starting frame $X_{t}$.

\subsection{3D Segmentation Transformation} \label{ssec:3DSegmentationTransformation}

Even though segmentation maps are two-dimensional objects, we propose to cast the segmentation prediction problem as a 3D problem since the underlying scene is three-dimensional. We first learn to predict the depth of the scene in unsupervised manner, which can also produce the ego-motion between consecutive frames of the video sequence. Recent work~\cite{mahjourian2018unsupervised, 8100183} has shown learning of depth and ego-motion without supervision and from monocular RGB videos only.    Using these predictions we can construct a 3D representation of the scene and use it to make predictions about future segmentation maps.

\textbf{Learning an unsupervised depth and ego-motion model.} We train the unsupervised depth and ego-motion model of \cite{mahjourian2018unsupervised} on the Cityscapes dataset using the open-source code, as it has shown higher accuracy in both depth and ego-motion estimates.  This method uses convolutional networks to estimate depth and ego-motion from raw RGB frames.  During training, depth and ego-motion estimates are used to reconstruct RGB images or point clouds from adjacent frames.  Loss functions on these reconstructions are used as supervision for learning depth and ego-motion.  The trained models can estimate depth from a single frame and ego-motion from a pair of frames.  While it is possible to train the depth and ego-motion model concurrently and end-to-end with the segmentation prediction model, for the sake of modularity and simplicity we chose to train these components separately.

\textbf{Applying the learned depth and ego-motion model.} Given a current frame $X_{t}$, we apply the trained depth model to produce an estimated depth map for the current frame, $D_{t}$. Subsequently, given $X_{t}$, and $X_{t-3}$, the frame three time steps in the past, we apply the trained ego-motion model to estimate the motion between frames $T_{t-3 \rightarrow t}$. Ego-motion is represented by a 6D vector describing the camera's movement as translation in x, y, z and rotation described by pitch, yaw, and roll.  While the rotation components are in radians, the translation components are in some scale-invariant unit chosen by the ego-motion model.  However, this unit is consistent with the distance unit learned in the depth model--since the two models are trained jointly--and therefore the predictions from these two models can be used together. 

\textbf{Obtaining 3D future segmentation.}
The camera intrinsics are then used to obtain a structured 3D point cloud $Q_t$ from the estimated depth map $D_{t}$:

\begin{equation}\label{eq:qt}
Q_t^{ij} = D_t^{ij} \cdot K^{-1} [i, j, 1]^T
\end{equation}
where K is the camera intrinsics matrix, and $[i, j, 1]$ are homogeneous coordinates iterating over all locations in the depth map.  We attach to each coordinate $i, j$ in $Q_t^{ij}$ the one-hot segmentation vector at the same coordinates in $\hat {S}_{t}^{ij}$.

Now given the prior camera ego-motion from previous frames e.g. until frame $t$, we can estimate or {\it predict} the future frame ego-motions, which are denoted as $\hat{T}_{t \rightarrow t+3}$, $\hat{T}_{t +3\rightarrow t+6}$, etc.
The next section (Section~\ref{ssec:Learn Future Ego-Motion}) describes how we learn to predict the future ego-motion vectors as a function of the prior trajectory, e.g. by using input ego-motion frames $T_{t-3j-3 \rightarrow t-3j}$, for $j \in \{0,1,2\}$. We experiment with this version later in the paper too, applying it to longer horizon predictions when prior frames are not available.

\begin{figure}[t]
  \includegraphics[width=\linewidth]{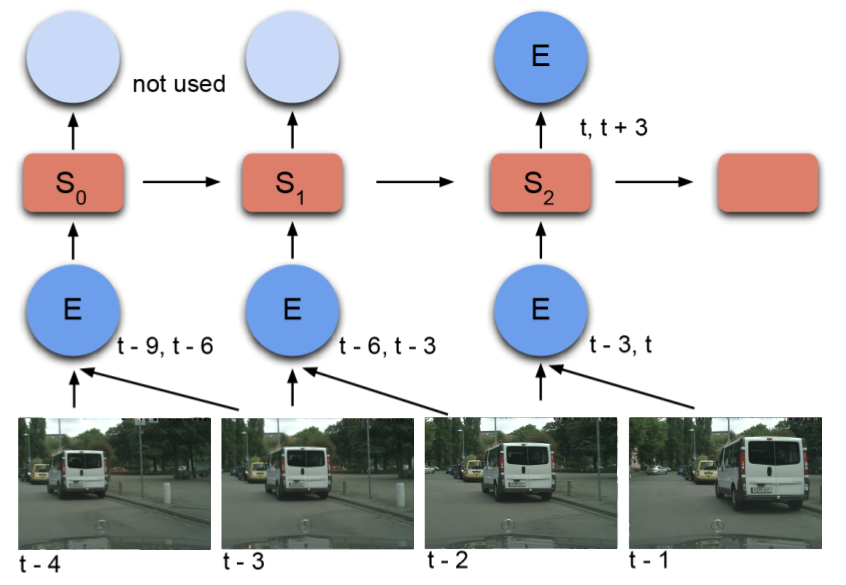} 
    \vspace{-4mm} 
  \caption{An LSTM model for future ego-motion predictions as a function of past ego-motions.}
  \label{fig:seq}    
\end{figure}

The future ego-motion prediction at the next step can then be used to generate a prediction for the scene point cloud in the future as:

\begin{equation}\label{eq:qt_3}
\hat Q^{ij}_{t+3} = \hat{T}_{t \rightarrow t+3} Q^{ij}_t
\end{equation}
where $\hat Q^{ij}_{t+3}$ denotes the predicted point cloud at frame $t+3$ with corresponding segmentation vectors at coordinates $i, j$.

We project the predicted point cloud to 2D space and use the attached segmentation vectors to construct a segmentation prediction for frame $t+3$ using the forward warp:
\begin{equation}\label{eq:r}
\hat{S}_{t+3}^{\hat{i}\hat{j}} = \hat{S}^{ij}_{t},
\end{equation}
where the coordinates $\hat{i}, \hat{j}$ are determined by

\begin{equation}\label{eq:ijhat}
[\hat{i}, \hat{j}, 1]^T = K \hat{Q}_{t+3}^{ij}.
\end{equation}

In order to predict more than one transformation (i.e. three frames) into the future, the above-mentioned transformation is applied recursively:
\begin{equation}\label{eq:qt_rec}
\hat Q_{t+3j+3} = \hat T_{t+3j \longrightarrow t+3j+3} \hat Q_{t+3j}
\end{equation}
where $j=1,2,3$, etc. and by using Eq.~\ref{eq:qt_3} for $j=0$.

\noindent Note that the future depth and ego-motion estimates are computed from current and past frames only.  Section~\ref{ssec:Learn Future Ego-Motion} introduces a method for predicting the future ego-motion trajectory. A similar approach can be applied for modeling individual object motions, which is not addressed here.

\begin{figure*}[t]
  \includegraphics[width=\linewidth]{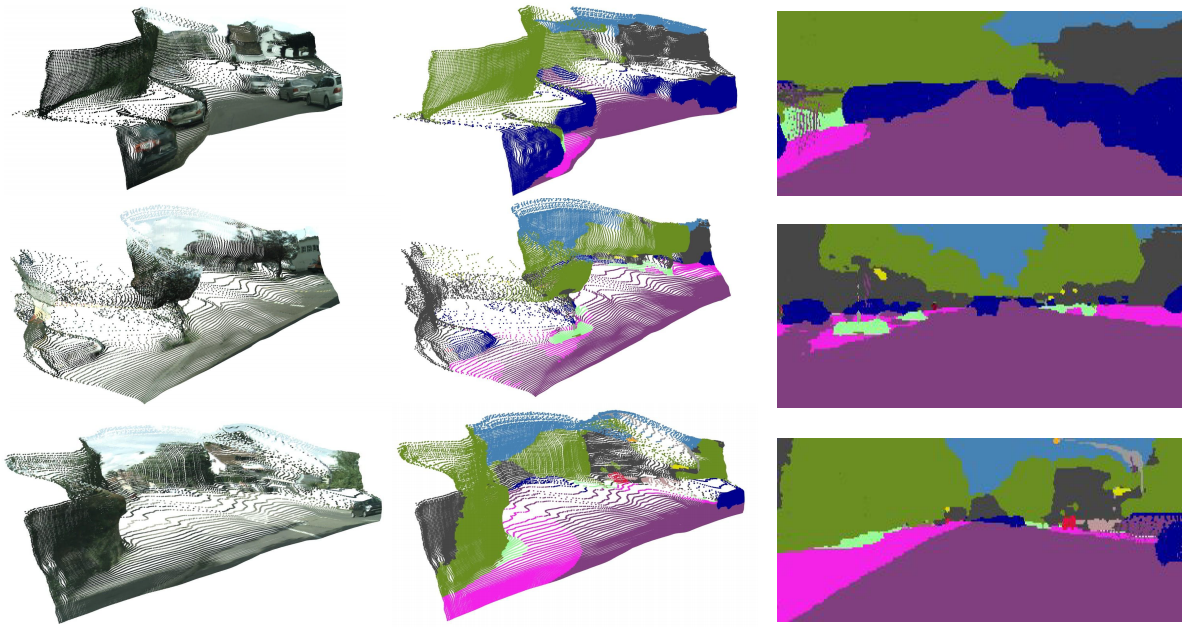}
    \vspace{-4mm} 
  \caption{The proposed 3D semantic segmentation approach considers the 3D scene for obtaining future frames.}
  \label{fig:seq_ptc}    
\end{figure*}

\subsection{Learning Future Ego-Motion} \label{ssec:Learn Future Ego-Motion}

In this section we describe how to learn the future ego-motion trajectory. 
As this work is grounded in a 3D geometric approach, we here treat the ego-motion as an SE3 transform (namely a rotation and a translation). We propose to learn the future ego-motion as a sequence of transforms which are a function of the prior ego-motion trajectory, Figure~\ref{fig:seq}. 
Our approach is in contrast to prior work~\cite{8237339} which learn a function directly from past scenes to the future scene and must develop the concept of motion from scratch. Instead, incorporating structure as SE3 transforms is more meaningful and leads to faster and more efficient training~\cite{Byravan17}.

Using our estimate of ego-motion from previous frames, e.g. $T_{t-3j-3 \longrightarrow t-3j}$, for $j \in \{0,1,2\}$, we can produce a prediction for future motion as:
\begin{equation}
\hat{T}_{t \rightarrow t+3} = \mu (T_{t-3 \rightarrow t}, T_{t-3 \rightarrow t}, T_{t-3 \rightarrow t}),
\end{equation}
\noindent where $\mu$ is a nonlinear function, to be learned from the observed motion sequences.
This motion prediction can then be used to generate a prediction for the scene point cloud in the future, as described in Section~\ref{ssec:3DSegmentationTransformation}.
For example, in our experiments, we use the prior ego-motions for $t-12$ to $t-9$, $t-9$ to $t-6$, $t-6$ to $t-3$, and $t-3$ to $t$, in order to predict the future ego-motion for $t$ to $t+3$, which is in accordance with prior work~\cite{8237339} which used 4 previous frames.

To estimate the unknown nonlinear function $\mu$ which should predict future ego-motion from previous ego-motion trajectory, we train a three-layer recurrent neural network for future ego-motion prediction. More specifically we use an LSTM network~\cite{lstm97} since the LSTM recurrent neural networks learn better relationships in time and offer stable training without vanishing or exploding gradients. In the LSTM network we design, each layer consists of a basic LSTM cell with six units, corresponding to the 6D ego-motion vector (with 3 values for rotation and 3 for translation). We use past give frames, $X_{t-12}$, $X_{t-9}$, $X_{t-6}$, $X_{t-3}$, and $X_{t}$ to generate three past network-estimated\footnote{The ego-motion vectors are previously learned in an unsupervised way from the monocular video and are not provided as ground truth in the data.} ego-motion vectors, $T_{t-12 \longrightarrow t-9}$, $T_{t-9 \longrightarrow t-6}$, $T_{t-6 \longrightarrow t-3}$, $T_{t-3 \rightarrow t}$. The output from the last layer is then considered the future ego-motion, $\hat T_{t \rightarrow t+3}$. The $l_1$ loss between the originally estimated future ego-motion $T_{t \rightarrow t+3}$ and the RNN-predicted future ego-motion $\hat T_{t \rightarrow t+3}$ is used for training:
\begin{equation}
\mathcal{L}_{\ell_1}(\hat T_{t \rightarrow t+3},T_{t \rightarrow t+3})=\sum_{i=1}^{6} \|T_{t \rightarrow t+3_{i}}-\hat T_{t \rightarrow t+3_{i}}\|
\end{equation}

We then use the predicted future ego-motion to produce future frame segmentations, and as seen in later experiments, predicting the future ego-motion is beneficial to the future segmentation.

\subsection{Inpainting Projected Segmentations}
\label{sec:inpaint}

Since some parts of the future scene may not be present in the starting frame (parts of the scene may have been occluded or are out of the field of view), transforming and projecting segmentation maps will yield missing pixels. This is undesirable for the method since the missing pixels create visual artifacts in predicted segmentation maps.  They also reduce the model's accuracy by failing to provide a prediction for some image coordinates.

To address this problem, an inpainting step is applied in the segmentation space after every motion transformation.
The inpainting is implemented by replacing every missing pixel with the most frequent segmentation class from its surrounding pixels.  More specifically, for each class in the segmentation map we compute a per-pixel neighbor count which reflects how many of the surrounding coordinates are ``on'' for that segmentation class.  The neighbor count is computed from $S$, the segmentation map generated by projecting the point cloud as

\begin{equation}
{N}^{ijk} = \sum_{j - 1}^{j + 1}\sum_{i - 1}^{i + 1} \hat{S}^{ijk}.
\end{equation}

We then select the most frequent surrounding class by computing a hardmax over $N$ at each coordinate $i, j$:

\begin{equation}
\text{Filler}^{ijk} = \text{hardmax}_k ({N}^{ijk}).
\end{equation}

The inpainted segmentation prediction $\mathbf{\hat{S}}$ is computed as:

\begin{equation}
\mathbf{\hat{S}}^{ijk} = 
     \begin{cases}
       \hat{S}^{ijk}, &\quad\text{if Eq. \ref{eq:r} yields a value at} (i, j)\\
       \text{Filler}^{ijk}, &\quad\text{otherwise}\\
     \end{cases}
\end{equation}

\textbf{Implementation details.} The model is trained about 1M steps. We used a learning rate decaying from $1e-4$ to $1e-6$, using momentum optimizer with momentum of $0.9$. Additional details are provided in the Supplementary material. 
\begin{figure*}[t]
  \includegraphics[width=\linewidth]{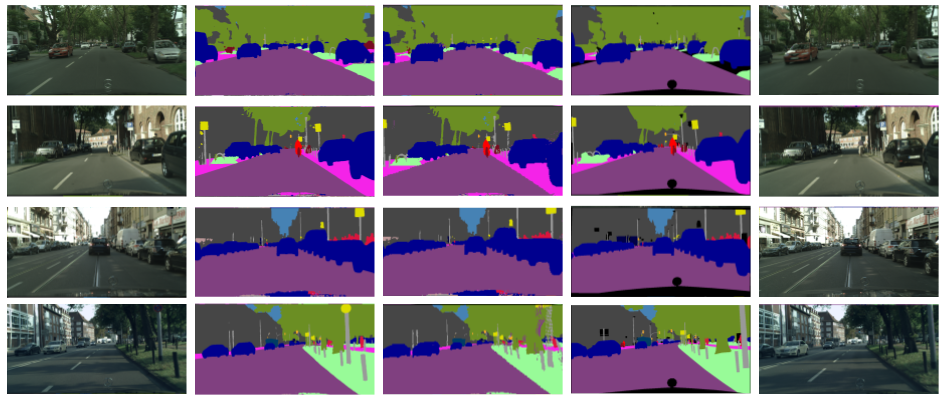} 
    \vspace{-4mm} 
  \caption{Results of the future 3D segmentation prediction model 3 frames ahead. From left to right: the RGB input frame (16), its corresponding segmentation, the predicted segmentation for frame 19, the ground truth segmentation of frame 19, and the RGB image of frame 19.}
  \label{fig:seq_schematic}    
\end{figure*}

\begin{figure*}[t]
  \includegraphics[width=\linewidth]{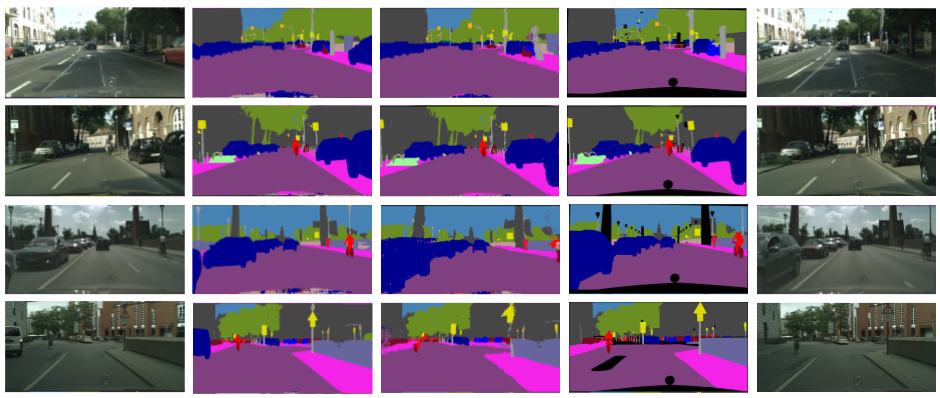} 
    \vspace{-4mm} 
  \caption{Future 3D segmentation prediction results for 9 frames ahead. For predictions of $10 \rightarrow 19$. From left to right: the RGB input frame (10), its segmentation (10), the predicted segmentation for frame 19, the ground truth segmentation (19), the RGB  for 19.}
  \label{fig:seq_schematic10}    
\end{figure*}

\section{Experiments}

\subsection{Dataset and Evaluation Metric}
We use the Cityscapes dataset~\cite{Cityscapes} which contains video sequences of city streets as captured by driving a car. Each video stream is 1.8 seconds long, contains 30 frames, with a human drawn ground truth segmentation mask available for the 19th frame. In total, 2,975 training, and 500 validation video sequences are included in the data set. 

The primary evaluation metric employed is IOU (intersection over union), calculated by summing the total number of pixels correctly classified between two segmentation masks, and dividing by the total number of pixels attempted to be classified. The results are reported as the average IOU mean across the full dataset of $500$ validation sequences (the Cityscapes dataset has no labeling on the test set).

The typical setup to evaluate prediction of the future segmentation, is to segment frames up until the 4th, 7th, 10th, 13th or 16th frame, and predict the segmentation for the 19-th. That is, prediction is done for 0.9 seconds to 0.18 seconds ahead of time, respectively. 
In accordance with prior work, the main experiments report the results of prediction from the 10th and 16th frames, since at least 3 prior steps are needed as input.
However our method allows for prediction farther in the future and we evaluate results in these settings when possible.

\subsection{Comparison to the state-of-the-art}

We first present the results of the proposed approach in comparison to the state-of-the-art (Table~\ref{tab:application2}).
We compare to previous state-of-the-art method by Luc et al. \cite{8237339} but not to Jin et al.\cite{8237857} as the latter work did not report numerical results for their predictive parsing network alone and included future and past inputs.
As seen in Table~\ref{tab:application2}, short term prediction of our method outperforms the prior work of Luc et al\cite{8237339} (which in turn outperforms other strong baseline methods like 2d warping), and our method achieves comparable performance accuracy at mid-term prediction. Visual results are presented in Figure \ref{fig:seq_schematic} for $16\rightarrow19$ and Figure \ref{fig:seq_schematic10} for $10\rightarrow19$ predictions. 

Furthermore, an advantage of our method is that the scene can be represented in 3D in the future time, as seen in Figure~\ref{fig:seq_ptc}. The main contribution of our approach is that we can start viewing segmentation, especially future segmentation, as a 3D problem rather than 2D as previously done.

Previous work also only reported up to 9 frames into the future. As such, we are not able to compare our longer term predictions to 
previous results. Subsequent results later in the paper show predictions further in the future.

\begin{table}
\caption{Future semantic segmentation results. IOU mean.}
\begin{center}
\label{tab:application2}
\begin{tabular}{|l|c|c|}
\hline
Method           & $10 \rightarrow 19$  & $16 \rightarrow 19$ \\ \hline
Baseline             & 0.3956   & 0.5461    \\ \hline
Luc et al.  &\textbf{0.4780}  & 0.5940   \\ \hline
Future segmentation in 3D (Ours) &0.4540  & \textbf{0.6147}    \\ \hline

\end{tabular}
\end{center}
\end{table}

We next test the results of individual components of our approach.

\subsection{3D Transformation Model}

\begin{table*}[ht]
\begin{center}
\caption{IOU mean results for future segmentation predictions for 3 - 15 frames into the future (i.e. 0.18 to 0.9 seconds). The future ego-motion model is a copy of prior ego-motion since, in some cases, not enough previous frames are available to estimate it.}
\vspace{2mm}
\scalebox{0.85}{
\label{tab:inpaintingXception}
\begin{tabular}{|c|c|c|c|c|c|}
\hline
Input Frame & Target Frame (GT) & \# Motion Transforms & Segmentation Copy (Baseline) & Motion Transform (Ours) & \begin{tabular}[c]{@{}l@{}}Motion Transform \\with Inpainting (Ours)\end{tabular} \\ \hline
4     & 19       & 5         & \textbf{0.3329}  & 0.2862        & 0.3289                                                                  \\ \hline
7     & 19       & 4         & 0.3595   & 0.3401        & \textbf{0.3866}                                                                  \\ \hline
10    & 19       & 3         & 0.3956   & 0.3982        & \textbf{0.4346}                                                                  \\ \hline
13    & 19       & 2         & 0.4507   & 0.4778        & \textbf{0.4993}                                                                  \\ \hline
16    & 19       & 1         & 0.5461   & 0.6077        & \textbf{0.6119}                                                                  \\ \hline
\end{tabular}
}
\end{center}
\end{table*}

We first test the improvements that were contributed by the main 3D transformation model (Motion transformation). 
More specifically, we test the approach with a simple egomotion-copy model, instead of learning it. The future ego-motion in that case is obtained by
using the estimate of camera motion from $t-3$ to $t$, and assuming the same motion for $t$ to $t+3$.
This motion prediction can then be used to generate a prediction for the scene point cloud in the future.   
While this is less accurate than learning it, as shown in the next section, this approach has the advantage that it can be applied with very few available prior frames. Thus we can predict and evaluate future frame segmentation for frames as far as 15 frames in advance in Cityscapes (prior work~\cite{8237339} only reports results for 3 and 9 future frames).

The results of the full pipeline, but with the ego-motion `copy' method, are reported in Table~\ref{tab:inpaintingXception}. The table shows results for predicting future frames starting from frame 4 i.e. $4 \rightarrow 19$ and up to $16 \rightarrow 19$.
It shows the baseline segmentation model and then in turn shows components of our algorithm, namely  of only applying 3D transformations in the scene, and transformations with inpainting. 

We observe improvements for future frame predictions, relative to the segmentation copy model baseline, which indicates the advantage of employing learned 3D structure for the task of future frame segmentation. Most notably the motion transformation model provides significant improvements, over models that do not transform the scene, especially so for mid-term and short-term predictions. Even so, it is a very simple estimate of the future ego-motion. The predictions for longer-term frames are less reliable as they are using a value that is farther in the past. We note that the motion transformation for the $4 \rightarrow 19$ is underperforming because the ego-motion function prediction, itself has only a single frame for computing the ego-motion and we obtained it by duplicating a frame due to a missing value, which is not realistic. For reference the $4 \rightarrow 19$  model will perform at $36.8$ IOU, if the estimated ego-motion is used, rather than the `copy' one applied multiple times.

\subsection{Inpainting}

\begin{figure}[t]
  \includegraphics[width=\linewidth]{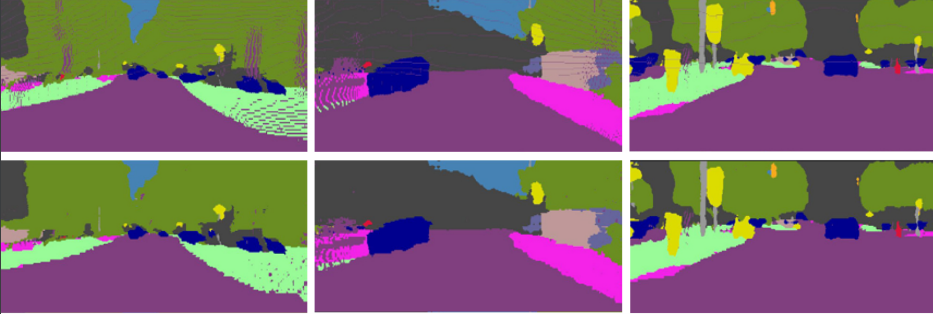} 
  \caption{Results showing predicted segmentations without (top) and with (bottom) inpainting. The examples show the results for different time spans; from left to right: 4{$\rightarrow$}19, 10{$\rightarrow$}19, 16{$\rightarrow$}19.}
  \label{fig:seq_inpaint}    
\end{figure}

The inpainting is also helpful, most notably so for longer-term predictions since more values are missing, the more transformations are done (Table~\ref{tab:inpaintingXception}). The inpainting is  performed after each motion transformation step which helps in removing of the accumulation of missing values. From these results we observe that if more reliable future predictions are available, more accurate future segmentations will occur and that the longer-term predictions are hurt by the fact that no prior ego-motion trajectories are available. With this approach we observe that using 3D geometry in learning is beneficial. 
Figure~\ref{fig:seq_inpaint} visualizes the effects of applying inpainting. Our results show consistent improvements with inpainting applied.  

\subsection{Future Ego-Motion Learning}

We here apply the learning of the future ego-motion trajectory, as described in Section~\ref{ssec:Learn Future Ego-Motion}. In the experimental setup, three prior ego-motion vectors are used as input to future ego-motion predictions. This is in accordance with prior work which use 4 prior frames as input. With that setup however, only mid- to short-term predictions, namely $10 \rightarrow 19$ and $16 \rightarrow 19$ can be evaluated. 

Table~\ref{tab:application} shows the results for mid- to short-term predictions with learning of future ego-motion. As seen, learning the future trajectory has positive effect by improving the future scene segmentation estimates further than the 3D transformation and inpainting model alone. 
Figures~\ref{fig:seq_schematic}, and ~\ref{fig:seq_schematic10} show example results where we can see that the model has been able to predict the scene as expected for the future frames. 

More importantly our method demonstrates that the 3D structure can be exploited effectively and that these components can be learned independently. Furthermore, in the case of future ego-motion prediction, it is being learned without additional supervision, with supervision provided from ego-motion estimates learned from an unsupervised model.
Our method shows that the application of 3D depth and ego-motion provides gains in future frame segmentation prediction and therefore provides an advantageous starting point for future semantic segmentation.

\begin{table}[ht]
\begin{center}
\caption{Learned Future Ego-Motion: IOU mean results for future segmentation predictions, when the future ego-motion vectors are learned from the history of prior trajectory.}
\vspace{2mm}
\scalebox{1.0}{
\label{tab:application}
\begin{tabular}{|l|c|c|}
\hline
Method           & $10 \rightarrow 19$  & $16 \rightarrow 19$ \\ \hline
Baseline             & 0.3956   & 0.5461    \\ \hline
Motion Transform & 0.3982  & 0.6077   \\ \hline
Motion Transform + Inpainting   &0.4346     & 0.6119   \\ \hline
Motion Transform + Inpainting & & \\ + Learning Future Ego-Motion  &\textbf{0.4540}  & \textbf{0.6147}    \\ \hline
\end{tabular}
}
\end{center}
\end{table}

\subsection{Error introspection}

\begin{figure}[t]
  \includegraphics[width=\linewidth]{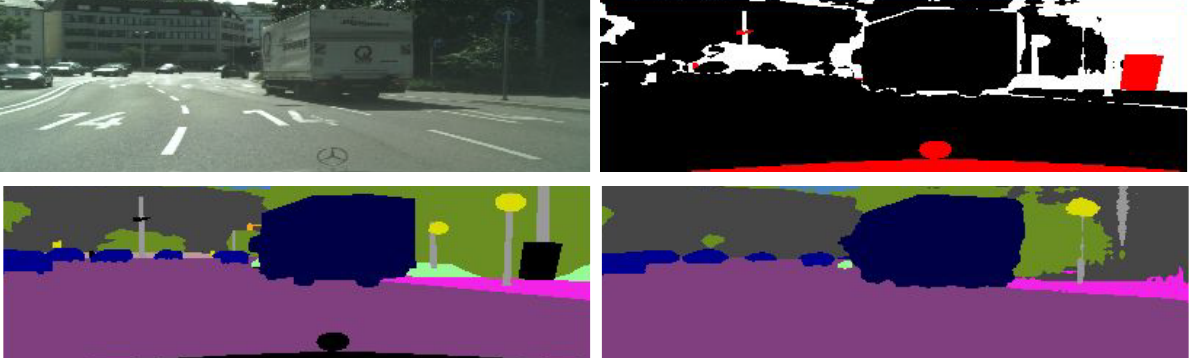}
  \includegraphics[width=\linewidth]{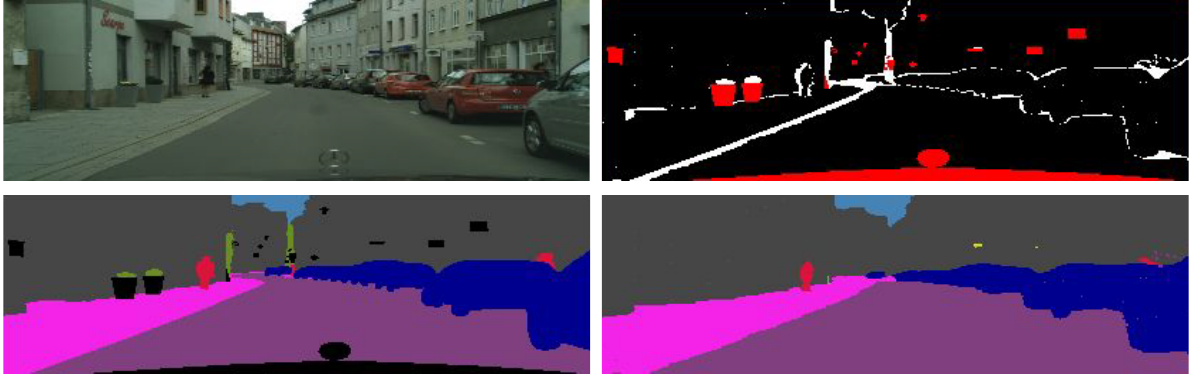} 
  \caption{Prediction errors of our model (top right image per panel.) The bottom rows are the ground truth and future predicted segmentations; errors are highlighted in white, red pixels are not considered.}
  \label{fig:seq_distance}    
\end{figure}

Figure~\ref{fig:seq_distance} further visualizes the errors of our method.
More specifically, the distances of the predicted model to the actual one is shown. 
Here we can see where the most errors occur. We can then observe that some errors are due to under-estimation of ego-motion or occlusions.

\section{Conclusions and Future Work}

We introduce a novel method for producing future frame segmentation masks, by employing the 3D structure of the scene, we learn the depth, ego-motion, and a model for future ego-motion trajectory prediction. An advantage of our framework is that it is combining 3D information, based on geometry components which are well understood, and machine learning components which can utilize the 3D information and learn from data. Moreover, we show competitive performance of future frame prediction, and can predict 0.9 seconds away from the starting frame.\par 

As future work, we anticipate exploration of more advanced techniques and strategies to fill larger areas of missing pixels, e.g. with GANs, will increase the quality of our predictions. Additionally, an extension to predict object trajectories is possible using the same future ego-motion prediction framework. This could improve the applicability of our work to a broader spectrum of scenarios.
We further plan to extend this work and demonstrate its effectiveness, by predicting future events for better motion planning, e.g. in the context of human-robot interaction.\par

{\small
\bibliographystyle{ieee}
\bibliography{ms.bbl}

\begin{thebibliography}{10}\itemsep=-1pt

\bibitem{Byravan17}
A.~Byravan and D.~Fox.
\newblock Se3-nets: Learning rigid body motion using deep neural networks.
\newblock {\em IEEE International Conference on Robotics and Automation
  (ICRA)}, 2017.

\bibitem{Chen2018-pv}
L.~C. Chen, G.~Papandreou, I.~Kokkinos, K.~Murphy, and A.~L. Yuille.
\newblock {DeepLab}: Semantic image segmentation with deep convolutional nets,
  atrous convolution, and fully connected {CRFs}.
\newblock {\em IEEE Trans. Pattern Anal. Mach. Intell.}, 40(4):834--848, Apr.
  2018.

\bibitem{journals/corr/Chollet16a}
F.~Chollet.
\newblock Xception: Deep learning with depthwise separable convolutions.
\newblock {\em CoRR}, abs/1610.02357, 2016.

\bibitem{Cityscapes}
M.~Cordts, M.~Omran, S.~Ramos, T.~Rehfeld, M.~Enzweiler, R.~Benenson,
  U.~Franke, S.~Roth, and B.~Schiele.
\newblock The cityscapes dataset for semantic urban scene understanding.
\newblock {\em CVPR}, 2016.

\bibitem{6909657}
D.~F. Fouhey and C.~L. Zitnick.
\newblock Predicting object dynamics in scenes.
\newblock In {\em 2014 IEEE Conference on Computer Vision and Pattern
  Recognition}, pages 2027--2034, June 2014.

\bibitem{7780459}
K.~He, X.~Zhang, S.~Ren, and J.~Sun.
\newblock Deep residual learning for image recognition.
\newblock In {\em 2016 IEEE Conference on Computer Vision and Pattern
  Recognition (CVPR)}, pages 770--778, June 2016.

\bibitem{6248012}
M.~Hoai and F.~D. la~Torre.
\newblock Max-margin early event detectors.
\newblock In {\em 2012 IEEE Conference on Computer Vision and Pattern
  Recognition}, pages 2863--2870, June 2012.

\bibitem{lstm97}
S.~Hochreiter and J.~Schmidhuber.
\newblock Long short-term memory.
\newblock {\em Neural Computation}, 1997.

\bibitem{8237857}
X.~Jin, X.~Li, H.~Xiao, X.~Shen, Z.~Lin, J.~Yang, Y.~Chen, J.~Dong, L.~Liu,
  Z.~Jie, J.~Feng, and S.~Yan.
\newblock Video scene parsing with predictive feature learning.
\newblock In {\em 2017 IEEE International Conference on Computer Vision
  (ICCV)}, pages 5581--5589, Oct 2017.

\bibitem{DBLP:journals/corr/KalchbrennerOSD16}
N.~Kalchbrenner, A.~van~den Oord, K.~Simonyan, I.~Danihelka, O.~Vinyals,
  A.~Graves, and K.~Kavukcuoglu.
\newblock Video pixel networks.
\newblock {\em CoRR}, 2016.

\bibitem{Kitani:2012:AF:2404742.2404759}
K.~M. Kitani, B.~D. Ziebart, J.~A. Bagnell, and M.~Hebert.
\newblock Activity forecasting.
\newblock In {\em Proceedings of the 12th European Conference on Computer
  Vision - Volume Part IV}, ECCV'12, pages 201--214, Berlin, Heidelberg, 2012.
  Springer-Verlag.

\bibitem{10.1007/978-3-319-10578-9_45}
T.~Lan, T.-C. Chen, and S.~Savarese.
\newblock A hierarchical representation for future action prediction.
\newblock In D.~Fleet, T.~Pajdla, B.~Schiele, and T.~Tuytelaars, editors, {\em
  Computer Vision -- ECCV 2014}, pages 689--704, Cham, 2014. Springer
  International Publishing.

\bibitem{8237339}
P.~Luc, N.~Neverova, C.~Couprie, J.~Verbeek, and Y.~LeCun.
\newblock Predicting deeper into the future of semantic segmentation.
\newblock In {\em 2017 IEEE International Conference on Computer Vision
  (ICCV)}, pages 648--657, Oct 2017.

\bibitem{7995953}
R.~Mahjourian, M.~Wicke, and A.~Angelova.
\newblock Geometry-based next frame prediction from monocular video.
\newblock In {\em 2017 IEEE Intelligent Vehicles Symposium (IV)}, pages
  1700--1707, June 2017.

\bibitem{mahjourian2018unsupervised}
R.~Mahjourian, M.~Wicke, and A.~Angelova.
\newblock Unsupervised learning of depth and ego-motion from monocular video
  using 3d geometric constraints.
\newblock In {\em The IEEE Conference on Computer Vision and Pattern
  Recognition (CVPR)}, June 2018.

\bibitem{DBLP:journals/corr/MathieuCL15}
M.~Mathieu, C.~Couprie, and Y.~LeCun.
\newblock Deep multi-scale video prediction beyond mean square error.
\newblock {\em ICLR}, 2016.

\bibitem{6126279}
M.~Pei, Y.~Jia, and S.~C. Zhu.
\newblock Parsing video events with goal inference and intent prediction.
\newblock In {\em 2011 International Conference on Computer Vision}, pages
  487--494, Nov 2011.

\bibitem{DBLP:journals/corr/RanzatoSBMCC14}
M.~Ranzato, A.~Szlam, J.~Bruna, M.~Mathieu, R.~Collobert, and S.~Chopra.
\newblock Video (language) modeling: a baseline for generative models of
  natural videos.
\newblock {\em CoRR}, abs/1412.6604, 2014.

\bibitem{DBLP:journals/corr/Shalev-ShwartzB16}
S.~Shalev{-}Shwartz, N.~Ben{-}Zrihem, A.~Cohen, and A.~Shashua.
\newblock Long-term planning by short-term prediction.
\newblock {\em CoRR}, abs/1602.01580, 2016.

\bibitem{DBLP:journals/corr/Shalev-ShwartzS16}
S.~Shalev{-}Shwartz and A.~Shashua.
\newblock On the sample complexity of end-to-end training vs. semantic
  abstraction training.
\newblock {\em CoRR}, abs/1604.06915, 2016.

\bibitem{DBLP:journals/corr/SrivastavaMS15}
N.~Srivastava, E.~Mansimov, and R.~Salakhutdinov.
\newblock Unsupervised learning of video representations using lstms.
\newblock {\em ICML}, abs/1502.04681, 2015.

\bibitem{Villegas17}
R.~Villegas, J.~Yang, S.~Hong, X.~Lin, and H.~Lee.
\newblock Decomposing motion and content for natural video sequence prediction.
\newblock {\em ICLR}, 2017.

\bibitem{8100183}
T.~Zhou, M.~Brown, N.~Snavely, and D.~G. Lowe.
\newblock Unsupervised learning of depth and ego-motion from video.
\newblock In {\em 2017 IEEE Conference on Computer Vision and Pattern
  Recognition (CVPR)}, pages 6612--6619, July 2017.

\end{thebibliography}
}

\end{document}


\title{Future Semantic Segmentation Using 3D Structure \\ Supplementary Material}

\onecolumn
\maketitle
\ifwacvfinal\thispagestyle{empty}\fi

\textbf{Supplemental Figures:}
Supplemental figures follow. Figure 1 visualizes future segmentation model results for the prediction of frame 19 from frame 10. Figure 2 shows future segmentation examples of frame 19 from frame 16. Figure 3 contains additional side-by-side examples of future frame segmentation predictions, starting from frame 10 and predicting frame 19, without and with the inpainting method applied. Figure 4 similarly displays four examples of future frame segmentation, without and with inpainting applied, starting from frame 16 and predicting frame 19. Supplemental results employ a DeepLab ResNet-101 baseline segmentation model, while main results in primary text employ a more accurate Xception model. Future prediction results are conceptually similar with both models; however supplemental results here with the weaker segmentation model, naturally amplify early segmentation errors.

\textbf{Additional training details:}
All models were implemented in TensorFlow. In order to obtain depth and ego-motion as ground truth used for our 3D future segmentation model, we train the depth and ego-motion as follows:
 the Adam optimizer was used with $\beta_1$ = 0.9, $\beta_2$ = 0.999, and learning rate $\alpha$ = 0.0002. We train the model for 20 epochs and use the model at the end of the 20th epoch. We trained our future ego-motion LSTM model using the Momentum Optimizer ($\beta$ = 0.9) and exponential decaying learning rate (initial learning rate $\alpha$ = 0.0005, decay rate $k$ = 0.9, final learning rate $\alpha$ = 0.0001).
 The model was trained over 160K steps, with batch size of 2. 

\begin{figure*}[h]
  
  \includegraphics[width=\linewidth]{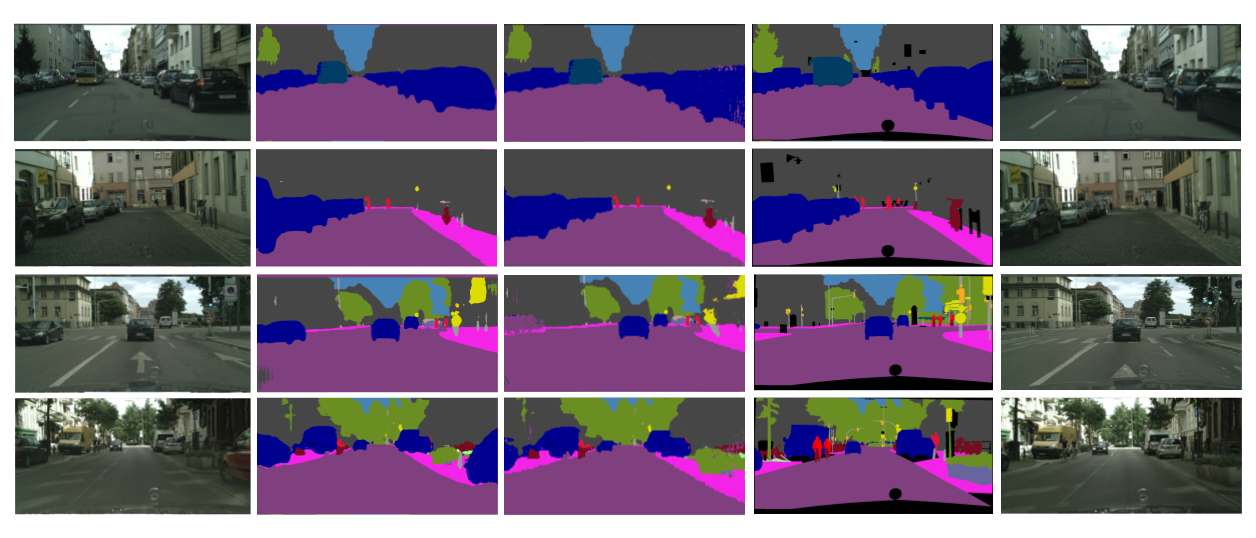}
    \vspace{-4mm} 
  \caption{Additional results of the 3D future segmentation prediction model. From left to right: the RGB input frame (10), its corresponding segmentation, the predicted segmentation for frame 19, the ground truth segmentation of frame 19, and the RGB image of frame 19. 
  } 
  \label{fig:seq_schematic_10_to_19}    
\end{figure*}

\begin{figure*}[h]
  \includegraphics[width=\linewidth]{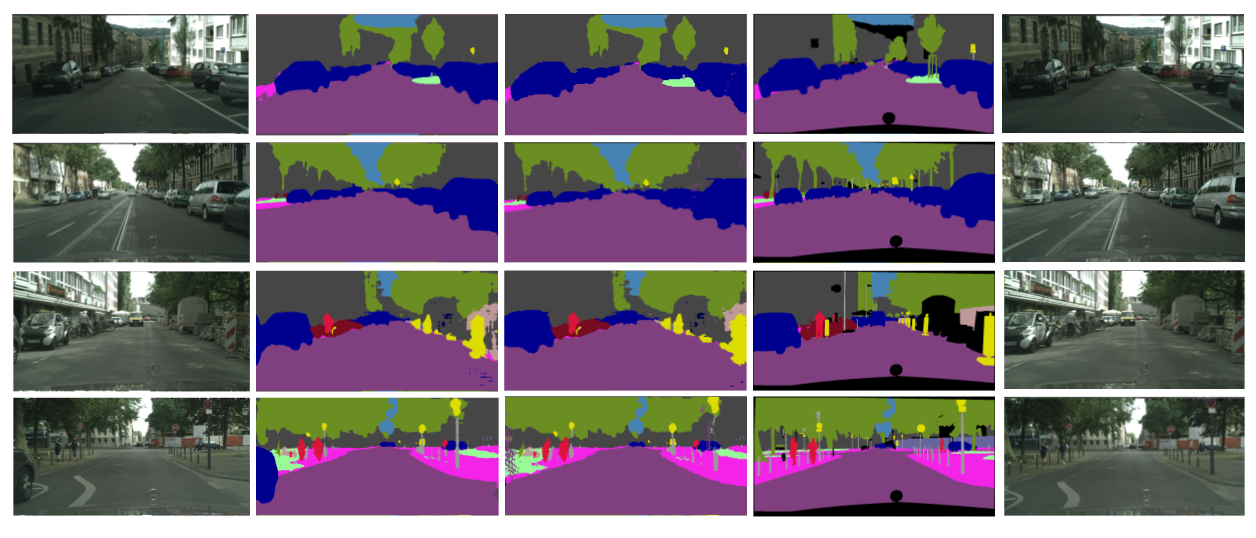}
    \vspace{-4mm}
  \caption{Additional results of the 3D future segmentation prediction model. From left to right: the RGB input frame (16), its corresponding segmentation, the predicted segmentation for frame 19, the ground truth segmentation of frame 19, and the RGB image of frame 19.}
  \label{fig:seq_schematic_16_to_19}
\end{figure*}

\begin{figure}[h]
  \includegraphics[width=\linewidth]{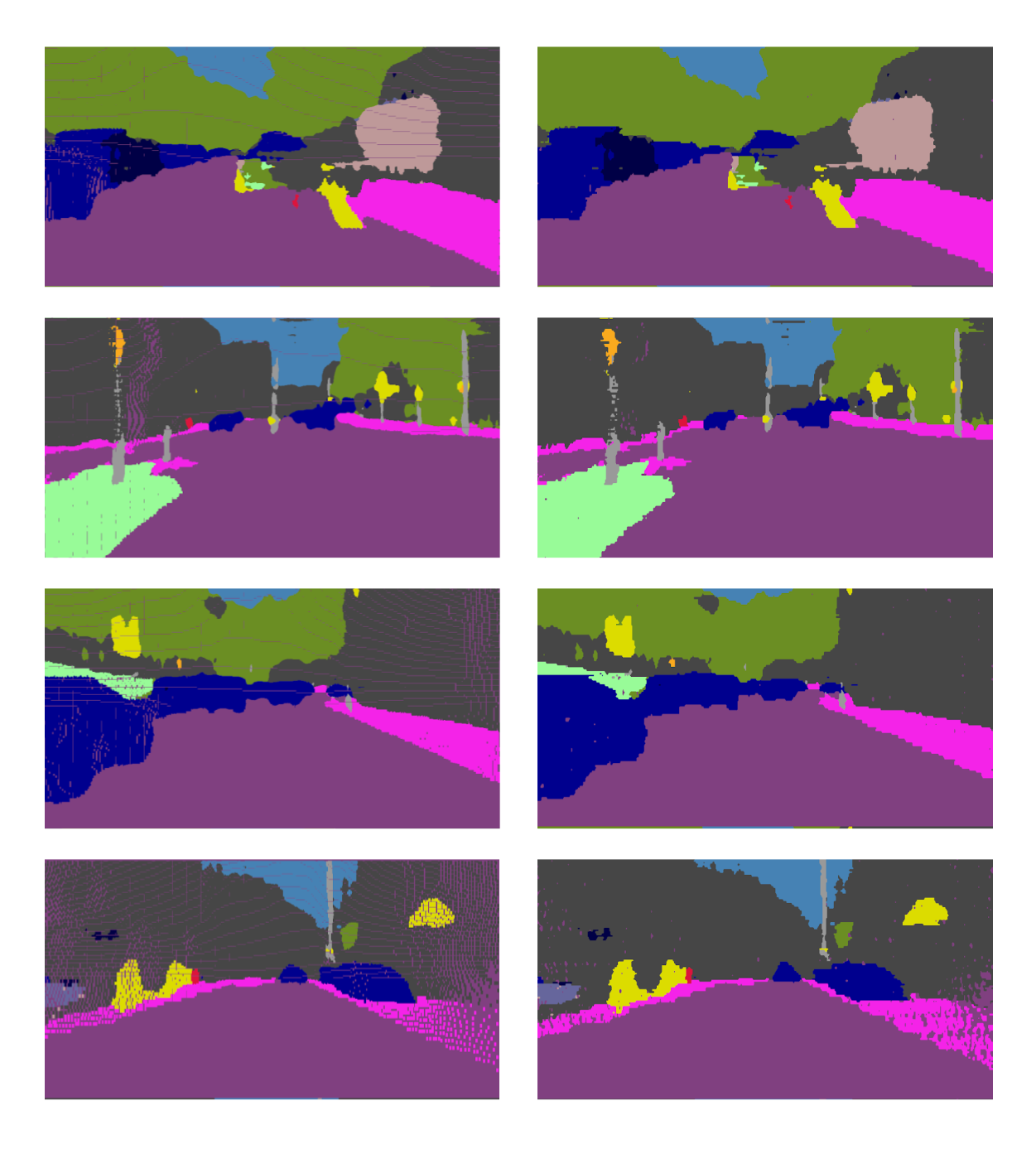} 
  \caption{Additional results showing predicted segmentations without (left) and with (right) inpainting. The examples show the results for 10{$\rightarrow$}19.}
  \label{fig:seq_inpaint_10to19}    
\end{figure}

\begin{figure}[h]
  \includegraphics[width=\linewidth]{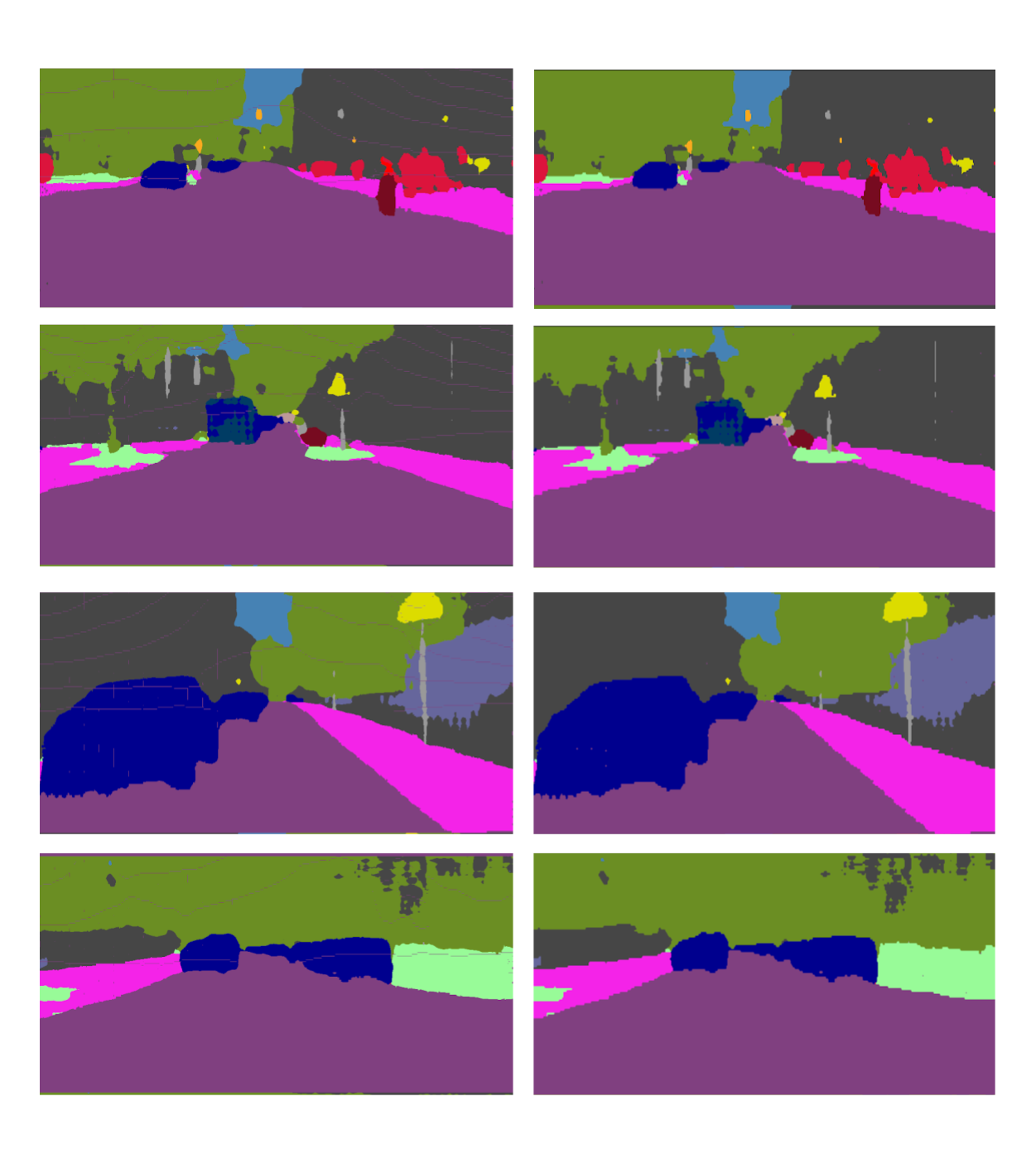} 
  \caption{Additional results showing predicted segmentations without (left) and with (right) inpainting. The examples show the results for 16{$\rightarrow$}19.}
  \label{fig:seq_inpaint_16to19}    
\end{figure}